# DATA AUGMENTATION AND TRANSFER LEARNING APPROACHES APPLIED TO FACIAL EXPRESSIONS RECOGNITION

Enrico Randellini, Leonardo Rigutini and Claudio Saccà

QuestIT Research Lab, Siena (Italy)

## ABSTRACT

*The face expression is the first thing we pay attention to when we want to understand a person's state of mind. Thus, the ability to recognize facial expressions in an automatic way is a very interesting research field. In this paper, because the small size of available training datasets, we propose a novel data augmentation technique that improves the performances in the recognition task. We apply geometrical transformations and build from scratch GAN models able to generate new synthetic images for each emotion type. Thus, on the augmented datasets we fine tune pretrained convolutional neural networks with different architectures. To measure the generalization ability of the models, we apply extra-database protocol approach, namely we train models on the augmented versions of training dataset and test them on two different databases. The combination of these techniques allows to reach average accuracy values of the order of 85% for the InceptionResNetV2 model.*

## 1. INTRODUCTION

The ability to build intelligent systems that accurately recognize the emotions felt by a person is an open challenge of Artificial Intelligence and undoubtedly represents one of the points of contact between the human and machine spheres. Since the face expression is the first thing we pay attention to when we want to understand a person's state of mind, facial expression analysis represents the first step in researching and building a human emotion classifier. In the facial expression recognition (FER) task, it is believed that there are six basic universal expressions, namely fear, sad, angry, disgust, surprise and happy [1]. To these emotions is often added a neutral expression.

Thanks to recent advances in the field of Machine Learning and Deep Learning, many FER systems have been proposed in the literature over the years, obtaining in some cases high accuracy values [2] [3]. On the other hand, greater levels of precision can be achieved taking into account the following issues:

1) it is observed a significant overlap between basic emotion classes [1] and differences in cultural manifestation of a given emotion [4];
2) the public image-labeled databases widely used to train and test FER systems may not be large enough;
3) the available datasets differ in the quality of pictures and how people express a given emotion. Some of databases are composed of images taken 'in the wild', where the labeled

    emotion is naturally manifested by the people while doing some action. This differs from other datasets where the pictures are taken when the people are posing with that particular expression;

4) the results strongly depend on the databases used for train and test the models. In the intra-database protocol, where train is carried out in one database and test in a subject independent set of the same database, the current methods achieve high accuracy, reaching around 95% [5] [6]. On the contrary, methods evaluated in the cross-database protocol, where train is carried out in one or more databases and the test in different databases, usually are obtained lower accuracy, ranging between 40% and 88% [5] [6] [7] [8].

An automated FER system can be seen as a supervised classification method comparing selected facial features from given image or video frame with faces within a database. It is a well established fact that computer vision tasks are optimally solved by convolutional neural network (CNN) and, it is usually necessary to have large databases in order to avoid overfitting [11][12][13]. Unfortunately, some public image-labeled databases used to train and test FER systems, such as Karolinska Directed Emotional Faces (KDEF)[14] and Extended Chon-Kanade (CK+)[15], are not sufficiently large. To overcome this problem were introduced data augmentation (DA) techniques. They are of two types: (i) geometric (e.g. rotation, translation and scaling) and color transformations that change the shape or the color of the starting images leaving unchanged their labels [16][7][8]; (ii) guided-augmentation methods (e.g. by generative adversarial network (GAN) [17]) that create new synthetic images with specific labels [18][8].

Another way to circumvent the obstacle of small train databases is making use of transfer learning and fine tuning. These are machine learning techniques enabling to use knowledge from previously learned tasks and apply them to newer, related ones [19][11].

Leveraging on the previous techniques of data augmentation and transfer learning, the recent work of Zavarez et al.[7] proposed a cross-database evaluation where a pre-trained VGG16 network is fine tuned on six databases, augmented by using geometrical transformations, and evaluated on a seven different database. Their test on CK+ database reaches an average accuracy of 88%. In a similar way, Porcu et al.[8], augmenting the train database KDEF with synthetics images by means of geometrical transformations and GAN techniques, reach an accuracy of 83% when a pre-trained VGG16 neural network is evaluated always in the CK+ test set.

The aim of this paper is to explore whether it is possible to further improve the accuracy and the ability to generalize on new data of automated FER systems. We will examine if the available data augmentation techniques allow us to enlarge the training datasets more than what has been done so far. Moreover, we will consider different CNN architectures in addition to the already used VGG16.

To address the issues related to the small size of KDEF database, we will make use of both DA techniques exposed above. We will apply geometric and color transformations in an offline mode storing the results as a new database. After that, we will build GAN models from scratch to generate novel synthetic images for every emotion. Moreover, in order to compare the results and enlarge the training dataset even further, we will make use of the synthetic images kindly made available by the group of Porcu et al.[8]. Our results will show that as the number of training data increases, will improve also the stability and performances of the models.

Inspired by the previous works [7] and [8], in this paper we will conduct both cross-database and intra-database protocol experiments. Once we have trained the models on the full KDEF dataset and its enlarged versions, we will evaluate them on the CK+ and JAFFE test set, showing a good ability to generalize on new data. Furthermore, we will apply a k-fold cross validation making use of a general database obtained by the union of the KDEF dataset, its augmented versions, and the CK+ and JAFFE datasets.

Encouraged by the remarkable results obtained in the field of image recognition, in this paper we will make use of transfer learning techniques applied to other CNN architectures not used before. In addition to the VGG16, we will consider the VGG19 [21], InceptionV3 [22] and InceptionResNetV2 [23] architectures already pre-trained on the ImageNet dataset [20]. Then, simply by modifying the finals layers of each models and fine tuning their values along the KDEF dataset and its augmented versions, we will be able to reach high accuracy values for the problem of face emotion recognition.

For example, we will show that the InceptionResNetV2 network applied to the CK+ dataset reaches a mean accuracy of 86.15% with a very close range variability. This is an index of excellent stability and generalization to new data.

The work is structured as follow. In section 2 we will describe the three datasets selected for the comparison (KDEF, CK+ and JAFFE) and as we pre-processed the images. In section 3, we will illustrate how we have increased the data by means of geometric transformations and GAN techniques and build the different train sets. In section 4, we describe how we have applied transfer learning and fine tuning techniques on pre-trained CNN. The section 5 describes the experimental setup and reports the results. We conduct our experiments by using Python 3.7.10, Tensorflow 2.4.1, and 12GB NVIDIA Tesla K80 GPU. Finally, in section 6, we present the conclusions and remarks.

## 2. DATA PREPARATION

### 2.1. Dataset

We conduct our analysis making use of three databases of images from subjects of different ethnicities, genders, and ages in a variety of environments: KDEF, CK+ and JAFFE databases. Their main properties are reported in the following:

1) KDEF: The Karolinska Directed Emotional Faces (KDEF) [14] consists of 4900 pictures of 70 subjects (35 males and 35 females), each of which has been photographed twice in each of the seven facial expressions at five different angles (full left profile, half left profile, straight, half right profile, full right profile). For our experiments we consider only the straight images with a total of 980 pictures.
2) CK+: The Extended Cohn-Kanade (CK+) [15] consists of 100 university students aged from 18 to 30 years. Each picture is a frame from videos where each subject was instructed to perform expressions that begin and end with the neutral expression. Once neglected the images belonging to the contempt expressions, which is not included in the list of considered emotions, we get a total of 902 pictures.
3) JAFFE: The JAFFE dataset [24] consists of 213 images of different facial expressions from 10 different Japanese female subjects. Each subject was asked to do seven facial expressions (six basic facial expressions and neutral).

In order to train a supervised classifier in the cross-database protocol, we take the KDEF as starting point to build the final training databases. Namely, on the KDEF we will apply various data augmentation techniques to obtain four different enlarged training databases. Finally, the models will be tested on the CK+ and JAFFE.

## 2.2. Face detection and image standardization

To understand what kind of emotion a person is feeling, we look at his eyes, if he wrinkles his nose, the shape of mouth and so on. All of these features manifest on the face of the person we are looking at, thus we have to concentrate only on the face, neglecting other parts of the body and the background. Our first action is to reduce all the images just to the rectangle containing the face. We run this transformation using the DNN module of the OpenCV library [25], with a confidence of 0.5 for face recognition.

We also fix a standard dimension for the input images, now containing only the portion of the face. We adopt (224, 224, 3), where the first two dimensions represent the number of the row and column pixels, while the third dimension is the number of colour channels in the RGB sequence. At this level we leave the pixel intensity between 0 and 255. As better explained in Section (4.3), we will change the normalization of the input values depending on the classifier model.

## 3. DATA AUGMENTATION

The KDEF is a small train database to solve a complex task of computer vision thus, in order to increase the amount of training data, we perform a Data Augmentation step. In particular, to generate new data from existing ones, we follow two approaches:

1) Geometrical and colour transformations;
2) Generation of synthetic images from scratch using GAN

## 3.1. Geometrical and colour transformations

In the first approach, we build a set of artificial synthetic images by modifying some geometrical and color characteristics of the original images. Namely, we define the following set of operators acting on the geometry and colors of each image leaving unchanged the expression of the face:

1) Random rotation: a function that rotates each image by a random factor $\rho$, namely a float which denotes the upper limit, as a fraction of $2\pi$, for clockwise and counterclockwise rotations. In the experiments we set $\rho = 0.1$.
2) Random zoom: a function which randomly zoom each image by a random factor $\zeta$. In the experiments we set $\zeta = 0.1$.
3) Random flip: a function which randomly flip each image on the horizontal mode.
4) Random height: a function which randomly adjusts the height by a random factor $\theta$, namely a positive float representing lower and upper bound for resizing the image vertically. In the experiments we set $\theta = 0.2$.
5) Random width: a function which randomly adjusts the width by a random factor $\omega$, namely a positive float representing lower and upper bound for resizing the image horizontally. In the experiments we set $\omega = 0.2$.
6) Random contrast: a function which randomly adjust the contrast of an image between $[1 - \gamma, 1 + \gamma]$. In the experiments we set $\gamma = 0.2$.

We apply the above transformations five times to each original image, obtaining five synthetic images with the same target emotion. In this way, we obtain a new dataset, called KDEF_DA_OL (standing for Data Augmentation Offline) made of 4900 new images. Finally, merging KDEF_DA_OL with the starting KDEF dataset, we get the first training dataset made of 5880 images. We call this dataset KDEF_OL.

## 3.2. Synthetic data generation: GAN

Introduced in 2014 [17], Generative Adversarial Network (GAN) are able to learn how to reproduce synthetic data that looks real. For example, computers can learn how to create realistic images and pictures of peoples that do not exist in reality. Generally, GANs train two neural networks simultaneously: the generator attempts to produce a realistic image to fool the discriminator, which tries to distinguish whether this image is from the training set or the generated set.

The authors of [8] used the GAN framework implemented for the DeepFake autoencoder architecture of the FaceSwap project (https://github.com/deepfakes/faceswap). Basically, the face images from the KDEF database are used as base to create novel synthetic images using the facial features of two images selected from the YouTube-Faces database [26]. The novel images differ between each other, in particular with respect to the eyes, nose and mouth, whose characteristics are taken from the two selected new images. The authors kindly shared their augmented database with 980 synthetic images that, for convenience, we call KDEF_GAN_PFA. Once standardized, we merge this set of images with the previous KDEF_OL dataset in order to obtain a larger dataset with 6860 images including the original KDEF and the augmented version with both offline and GAN techniques. We call this dataset KDEF_PFA.

In this work we apply GANs models for data augmentation in a different way than [8]. First, we group the pictures with the same expression of the KDEF database. To each group we add four images of famous actors with the same expression taken from the web. For example, the pictures in Figure 1 with a manifestly happy expression, once standardized, have been added to the 140 happy images of KDEF dataset. Thus, for each emotion we obtain a set of 144 pictures that will be used to train a couple of GANs generator and discriminator networks in order to generate new synthetic images sharing the same facial expression. We believe that this procedure introduces a certain variability into the train dataset, thus the produced synthetic faces will slightly differ from the parent faces in terms of pose, brightness and background.

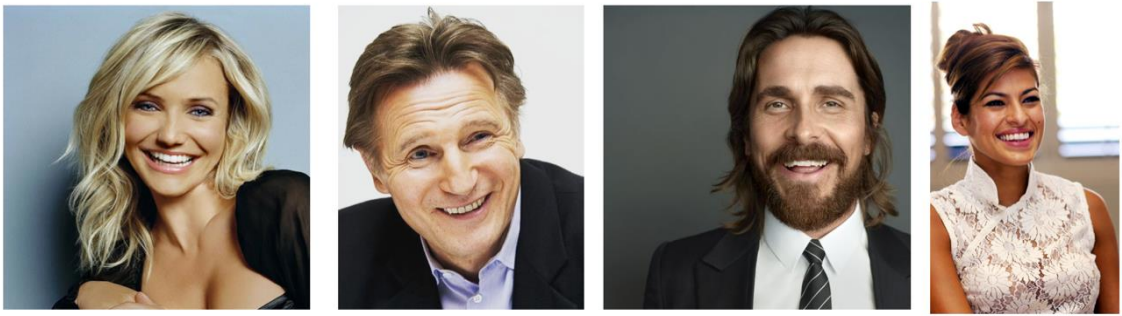

Figure 1. Pictures of known actors with a happy facial expression taken from the web

We assemble the discriminator model as a typical image classifier. In agreement with the structure of deep convolutional GAN (DCGAN) [27], as represented in Figure 2 the discriminator is a network made of a first convolutional 2D layer followed by five convolutional layers with striding to downscale the image by a factor of two every step. The result goes through flatten layer, followed by a dense sigmoid layer which returns a single output probability to classify the input picture as real or fake. In each striding layer we use a LeakyReLU activation function and a number of filters starting from 32 and doubling at each layer.

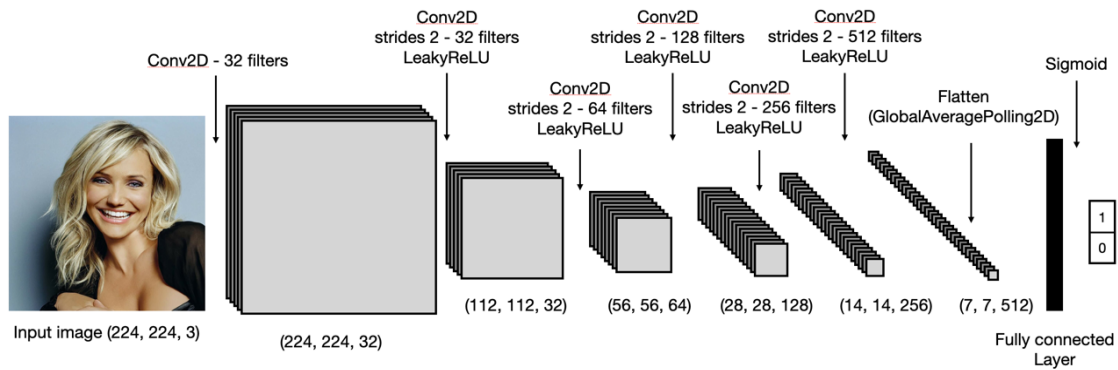

Figure 2. Discriminator architecture

As shown in Figure 3, the generator takes in input a noise vector from the latent dimension, chosen equal to 100, and generates an image. The network first upsamples the noise vector with a dense layer in order to have enough values to reshape into the first generator block. Each block consists of a transposed convolution 2D layer to upsample the image by a factor of two. We use 5 decoder blocks with LeakyReLU activation function and a final convolution 2D layer with hyperbolic tangent activation function to get a 3D tensor with the desired shape (224, 224, 3), which represents the final produced image.

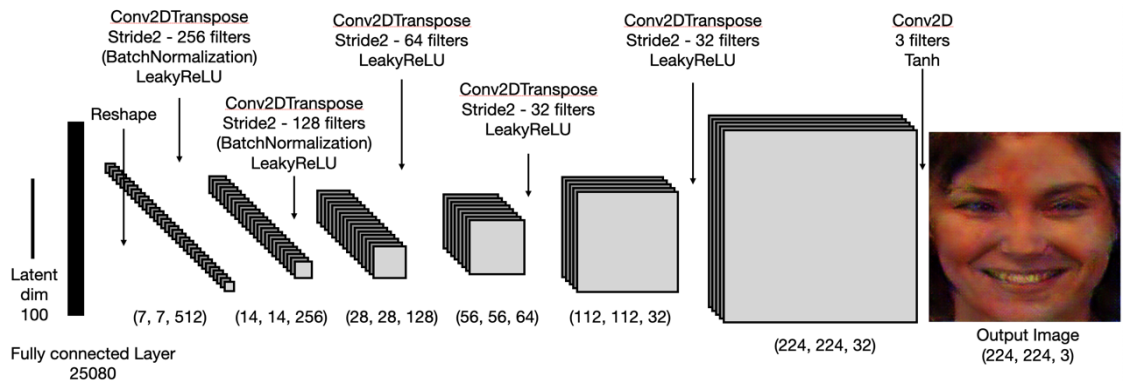

Figure 3. Generator architecture

We build the joined DCGAN by adding the discriminator on the top of the generator and train it applying the following steps. First, we send random noise to the generator, adding the output with real images to initially train only the discriminator. Then we freeze the discriminator and train the generator with the purpose to fooling the discriminator. We repeat this process iteratively for 2000 epochs using the Adam optimizer with a learning rate of 0.0002 and $\beta_1 = 0.5$ [28], and monitoring the quality of the generated images every 100 epochs. We start saving trained models starting with 1000-th epoch in order to use the most stable version depending on the quality of the produced images.

We repeat this procedure for each emotion. First, we train a DCGAN model, thus, using the most stable version, we generate 150 synthetic images. At the end we get with a total of 1050 fake images composing the dataset KDEF_GAN_Q. As before, we merge this dataset together the previous KDEF_OL in order to obtain another larger dataset with 6930 images including the original one and the augmented version with both offline and a second GAN technique. We called this dataset KDEF_Q.

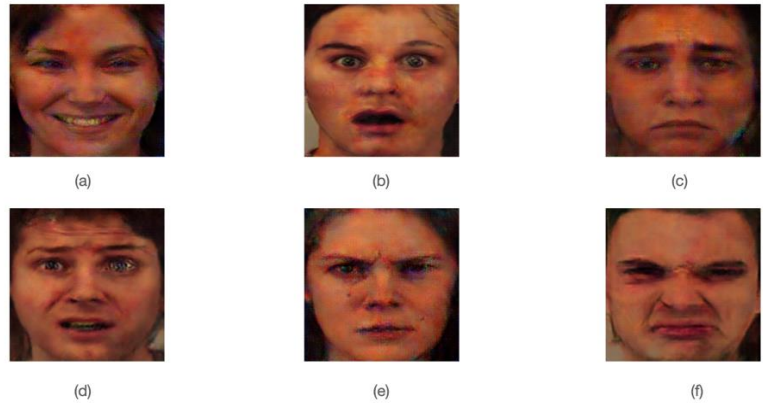

Figure 4. Examples of generated images by DCGAN models showing different face emotions:
(a) happy, (b) surprised, (c) sad, (d) afraid, (e) angry, (f) disgusted

We add the following remarks. Since the activation function of the last layer of the generator applies the hyperbolic tangent, we rescale the input images between -1 and 1 before training the DCGAN model. Thus, we subsequently rescale once again the generated images between 0 and 255, in agreement with the adopted standardization. Furthermore, we note that another stable configuration for generator model takes a batch normalization layer after the first two transposed convolution layers and, at the same time, a global average pooling 2D layer instead of the flatten one at the end of the discriminator model. In both cases the quality of the produced images is quite sufficient. As shown in Figure 4, we get images of people whose features clearly express typical facial emotions. Although the quality of the generated images it is not comparable with the latest GAN techniques as [29][30], we test these images with an emotion classifier in order to check if the predicted emotions coincide with those of the images. We refer the discussion of this test to Section 5.3.

In addition to the three previous dataset we also consider their union, namely a dataset containing the original KDEF, the augmented offline version, the augmented GAN version obtained by [8] and our augmented GAN version. This dataset contains 7910 images and has been called KDEF_PFA_Q. Summarizing, the datasets on which we will train and test the emotion classifiers are listed in Table 1.

Table 1. Main features of train and test dataset

| Dataset | Images | Usage |
|---|---|---|
| KDEF_OL | 5880 | Train |
| KDEF_PFA | 6860 | Train |
| KDEF_Q | 6930 | Train |
| KDEF_PFA_Q | 7910 | Train |
| CK+ | 902 | Test |
| JAFFE | 213 | Test |

## 4. MODELS AND TRAINING ALGORITHM

In this work we fine tune pre-trained models applying the transfer learning technique. We consider four deep learning architectures, each of which has placed a milestone in the problem of image recognition, namely the VGG16, VGG19 [21], InceptionV3 [31][22] and InceptionResNetV2 [32][23]. The models where trained on the ImageNet ILSVRC-2012 dataset (http://image-net. org/challenges/LSVRC/2012/), which includes more than one million images distributed along one thousand different classes [11].

The idea behind transfer learning consists in reusing the knowledge learned in solving a given problem and transferring it to solve a different but similar one [19] [11]. We consider deep networks already trained in a very big dataset to solve a problem of image classification. We thus reuse this knowledge, namely the values of the weights of the networks, to solve the problem of emotions classification. Furthermore, it is known that each layer of a neural network learns how to identify the features that are necessary to perform the final classification. Usually, lower layers identify lower-order features such as colors and edges, and higher layers compose these lower-order features into higher order ones such as shapes or objects. Hence, the intermediate layer has the capability to extract important features from an image which are useful for making a different kind of classification. Thus, to specialize the networks to our task, we applied fine-tuning technique freezing the values of the weights of a first part of the layers, and training the second part on ours specific dataset.

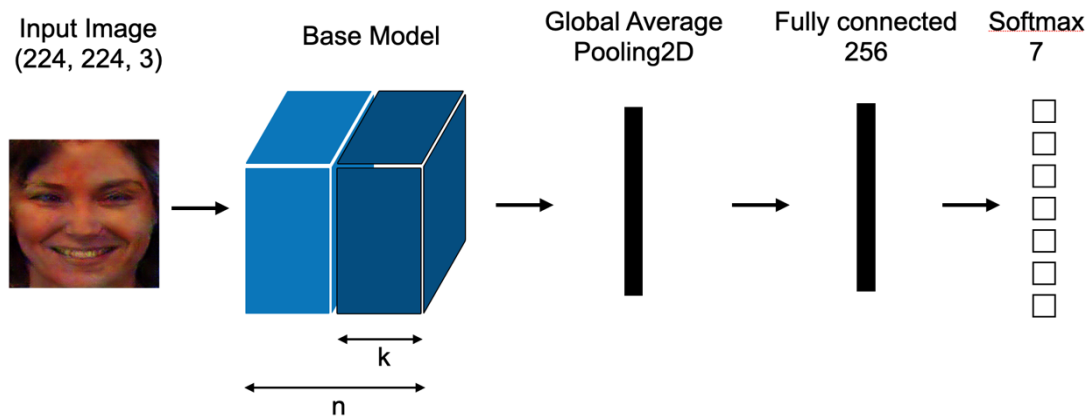

Figure 5. Architecture of neural networks used for transfer learning and fine tuning techniques. In the Base Model block, *n* represent the total number of layers, while *k* is the number of trainable layers during the fine tuning procedure.

### 4.1. Architecture of the models

We build the models joining the components one after the other. As shown in Figure 5, the first block is the base model, namely one of the pretrained networks. Tensorflow allows to download the base architecture where the weights are already trained in the ImageNet dataset. Since we want to fine tune the models in different datasets respect to the ImageNet one and with a different number of target classes, we specified the clause *include top=False* when downloading the model in order to remove the last layers related to the ImageNet classification task. Thus, we applied a global average pooling 2D layer, then a fully connected layer with 256 neurons and, lastly, a softmax activation function with seven outputs, corresponding to the seven possible emotions.

### 4.2. Fine tuning procedure

We divide the training procedure in two steps. Looking at the Figure 5, first we freeze the values of the weights of all the *n* layers of the base model, training for 10 epochs only the weights of the fully connected layer and the softmax function at the output. In this stage, the models are trained using the Adam optimizer with a learning rate equal to $10^{-3}$. Subsequently, we unlock the last *k* layers of the base model, allowing to tune their weights on the training datasets for 65 epochs. In this stage, we always use the Adam optimizer but with a learning rate equal to $10^{-4}$ to not move too far from the optimal position.

As shown in Table 2, each base model has a different number of layers, thus a different number of parameters. To find the optimal value of tuned layers *k* in the second step, we carry out some train and validation test on the training datasets. Thus, for each of model, we choose the value of *k* maximizing the validation accuracy.

Table 2. Features of the base models

| Base model | Layers | Tuned Layers | Params | Trainable params |
|---|---|---|---|---|
| VGG16 | 19 | 5 | 14714688 | 7079424 |
| VGG19 | 22 | 9 | 20024384 | 14158848 |
| InceptionV3 | 311 | 140 | 21802784 | 16215936 |
| InceptionResNetV2 | 780 | 371 | 54336736 | 40442464 |

### 4.3. Images normalization

The different base models were pre-trained on the ImageNet dataset using different normalizations for the input images. We must therefore adapt our datasets to these normalizations. Thus, for the InceptionV3 and InceptionResNetV2 models, we scale the pixel intensity between 0 and 1. Instead, for the VGG16 and VGG19 models, first we convert the input images from RGB to BGR, then we zero-center each color channel with respect to the ImageNet mean, namely $(103.939, 116.779, 123.68)$, without scaling.

## 5. EXPERIMENTAL SETUP AND RESULTS

We perform experiments following two strategies: a (i) *cross-datasets* approach, in which the model is trained and tested using different datasets, and (ii) *intra-datasets* approach, in which a global dataset is created by the union of the specific datasets and it is used for training and test.

### 5.1. Cross-datasets test

We implement the cross-datasets procedure by training the models on the datasets KDEF_OL, KDEF_PFA, KDEF_Q and KDEF_PFA_Q, then testing them on the CK+ and JAFFE datasets. To reduce the influence of random weights initialization, each architecture was trained and tested ten times. Thus, for each metric, we evaluate mean and standard deviation. The results for the accuracy values are summarized in Table 3.

Table 3. Mean accuracy and standard deviation of the 10 runs for each model
on the CK+ and JAFFE databases

| Train Dataset | Model | CK+ (%) | JAFFE (%) |
|---|---|---|---|
| KDEF_OL | VGG16 | 72.96 ± 8.55 | 42.72 ± 3.83 |
| | VGG19 | 80.99 ± 6.99 | 39.12 ± 5.52 |
| | InceptionV3 | 58.73 ± 8.07 | 42.66 ± 3.32 |
| | InceptionResNetV2 | 81.28 ± 4.65 | 39.23 ± 4.15 |
| KDEF_PFA | VGG16 | 68.44 ± 6.51 | 45.26 ± 3.53 |
| | VGG19 | 66.34 ± 8.65 | 45.12 ± 4.12 |
| | InceptionV3 | 50.60 ± 7.30 | 44.20 ± 2.56 |
| | InceptionResNetV2 | 78.73 ± 6.71 | 45.21 ±3.09 |
| KDEF_Q | VGG16 | 74.29 ± 5.59 | 40.61 ± 4.98 |
| | VGG19 | 81.54 ± 4.47 | 37.15 ± 2.91 |
| | InceptionV3 | 69.6 ± 8.37 | 43.19 ± 4.01 |
| | InceptionResNetV2 | **86.15 ± 3.54** | 42.58 ± 3.86 |
| KDEF_PFA_Q | VGG16 | 72.66 ± 5.40 | 46.10 ± 3.42 |
| | VGG19 | 71.93 ± 5.31 | 42.91 ± 7.60 |
| | InceptionV3 | 55.88 ± 5.74 | **47.56 ± 2.41** |
| | InceptionResNetV2 | 79.76 ± 4.53 | 44.84 ± 4.11 |

The results vary significantly between the two test databases. The ability of the models to generalize on new images is quite high on the CK+, while it lowers considerably on the JAFFE. This fact shows the importance to test the models in at least two different databases to measure their generalization ability in the cross-database protocol. Furthermore, we observe that the generalization ability is also conditioned by the similarity between the test and train datasets. Actually, we cannot ignore that the JAFFE dataset is highly biased in term of gender and ethnicity, namely it comprises only Japanese female subjects.

The tests performed on the CK+ set show that the InceptionResNetV2 architecture not only achieves the highest accuracy on the KDEF_Q train set, with a mean value of 86.15% and a max peak of 90.35% between the ten runs, but it is also the most stable model because the smallest range of variation. On the other way, the InceptionV3 model fails to generalize. The results change slightly in the case of the JAFFE test set. In this case the InceptionV3 is the best model on each train datasets and reaches the maximum value for the accuracy when is trained on the dataset KDEF_PFA_Q.

Table 4. Comparison of the accuracy values between our InceptionResNetV2 model trained on the
KDEF_Q database and other models tested on the CK+ database

| Method | Training Dataset | Accuracy (%) |
|---|---|---|
| Proposed | Augmented KDEF | 86.15 |
| Porcu et al.[8] | Augmented KDEF | 83.30 |
| Zavarez et al. [7] | Six databases | 88.58 |
| Hasani et al.[33] | MMI + FERA | 73.91 |
| Lekdioui et al.[34] | KDEF | 78.85 |

Computer Science & Information Technology (CS & IT) 159

Table 5. Comparison of the accuracy values between our InceptionV3 model trained on the KDEF PFA Q database and other models tested on the JAFFE database

| Method | Training Dataset | Accuracy (%) |
|---|---|---|
| Proposed | Augmented KDEF | 47.56 |
| Zavarez et al. [7] | Six databases | 44.32 |
| Ali et al.[33] | RaFD | 48.67 |
| Da Silva et al.[4] | CK | 42.30 |

In Table 4 and Table 5, we compare the accuracy achieved by the proposed best architectures with those achieved by state of the art cross-database experiments conducted for FER systems and tested, respectively, on the CK+ and JAFFE database. In the case of the CK+ test set, our result is second only to the approach proposed by Zavarev et al.[7], that trained a VGG16 model in a dataset composed by six different database enlarged by using geometrical and colour transformation, therefore on a number of images higher than ours. On the other hand, we can see that our result slightly improves that obtained by Porcu et al.[8]. Actually, our approaches are quite similar. They differ in the way we applied the GAN techniques to increase the number of images of the KDEF dataset and, of course, the model architectures. Namely their analysis uses only a VGG16 model.

Table 6. Precision and recall values of the best InceptionResnetV2 model trained on the KDEF_Q dataset and applied to the CK+ dataset

| Emotion | Precision (%) | Recall (%) |
|---|---|---|
| Angry | 89 | 38 |
| Disgust | 72 | 98 |
| Fear | 55 | 48 |
| Happy | 100 | 93 |
| Neutral | 96 | 96 |
| Sad | 48 | 71 |
| Surprise | 92 | 93 |

Finally, we present the values of the metrics precision and recall computed for the single emotion classes when considering the best combination of model architecture. In Table 6 we show the results obtained for the InceptionResNetV2 model trained on the KDEF Q dataset and tested on the CK+ for which we get a peak of accuracy of 90.35%. We remember that the precision for a given class measures the number of correctly predicted samples out of all predicted samples in that class. Instead, the recall for a given class measures the number of correctly predicted samples out of the number of actual samples belonging to that class. Thus, although the InceptionResnetV2 applied to CK+ is a good classifier, it has a high false positives rate on fear and sad faces because roughly half of samples predicted as fear or sad actually do not belong to these classes. At the same time, it struggles to recognize angry and fear faces because only the 38% and 48%, respectively, of angry and fear faces are correctly classified. As happened for the accuracy, the precision and recall values of the best model evaluated on the JAFFE dataset get drastically worse. This is because the training datasets differ greatly from the JAFFE one, which containing only Japanese female subjects. Thus, the results obtained in this case are not of great relevance and it is useless to show them.

### 5.2. Intra-datasets test

The last experiment that we conduct is made on the union of all the datasets consider so far, namely, referring to Section 2, the original KDEF dataset, its augmented versions KDEF_DA_OL, KDEF_GAN_PFA, KDEF_GAN_Q and, finally, the CK+ and JAFFE databases. In this dataset, made of 9025 images, we apply a k-fold cross validation with k equal to 5. Using the same training procedure and the same model architectures of the previous section, we get the results showed in Table (7) for the accuracy on the validation folds.

Table 7. Mean accuracy and standard deviation on the five validation folds

| Model | Accuracy (%) |
|---|---|
| VGG16 | 85.00 ± 1.83 |
| VGG19 | 97.61 ± 0.58 |
| InceptionV3 | 97.49 ± 1.83 |
| InceptionResNetV2 | 97.99 ± 1.07 |

As we expected, the accuracy values obtained in the intra-database protocol are greater then those obtained in the extra-database protocol. Once again, the InceptionResNetV2 remains the model with the greatest accuracy with a mean of 97.99%. We also note that the VGG16 models is unable to reach the same accuracy values as the other models.

### 5.3. Quality test for GAN generated images

Table 8. Accuracy values for the best VGG19 and InceptionResnetV2 model trained on KDEF_OL dataset and tested on each separated group of 150 generated fake images

| Emotion | VGG19 (%) | InceptionResnetV2 (%) |
|---|---|---|
| Angry | 91 | 100 |
| Disgust | 87 | 88 |
| Fear | 97 | 85 |
| Happy | 95 | 97 |
| Neutral | 93 | 98 |
| Sad | 96 | 36 |
| Surprise | 97 | 99 |

In Section 3.2 we generated 150 fake images for each emotion by means of GAN techniques. In order to test their quality, namely to check if each group of images is actually classified as belonging to that class, we use the best emotion classifiers that we found in Section 5.1. In particular, looking to Table 7, we consider the models VGG19 and InceptionResNetV2 with the highest accuracy and trained in the dataset KDEF_OL because it does not contain the generated fake images that we want to test. The results are shown in Table 8. As learned in Section 5.1, the classifiers struggle to recognize sad faces, in fact only the 36% of them are correctly recognized by the InceptionResnetV2 model. In all other cases the generated fake images are correctly classified with accuracy values higher the 85%, thus their quality is quite good for artificially increasing the original KDEF dataset.

### 6. CONCLUSIONS

In this paper we extensively investigated various techniques in order to build efficient supervised systems able to recognize human face expressions. The main obstacle to solve this task is the small size of the available training datasets. To get around this problem we made use of data augmentation techniques, such as geometrical transformations and training from scratch GAN models.

The experiments conducted in the cross-database protocol showed that the pretrained InceptionResnetV2 network, once fine tuned on an enlarged version of the KDEF database and tested on the CK+ test set, reaches a mean accuracy value of 86.15% with a close range of variation. Although the high values achieved for the accuracy, the model seem to suffer in recognizing emotions like fear and sad, for which we have obtained values of precision and recall under 70%. This problem is quite common to other FER systems and is probably related to the shape of faces that sometimes is very similar for these types of emotions.

The main obstacle to further increase the performances of the models and their ability to disentangle the recognition of face emotions remains the size and the composition of the training datasets. We showed that, even with few images for the training phase, GAN model can be built from scratch to obtain new synthetic images which are undoubtedly useful for obtaining the final performing FER systems. On the other hand, these synthetic images are of poor quality and all share the same appearance. In a future work, for the data augmentation step, we will apply transfer learning and fine tuning techniques to pretrained GAN models [36][37]. Thus, also with limited data, we will specialize GAN architectures to generate a large number of high quality images for each emotion type to be used, later, in the task of training an emotion recognizer.